\RequirePackage{amsthm}

\documentclass[pdflatex,sn-mathphys-num]{sn-jnl}

\usepackage{graphicx}%
\usepackage{multirow}%
\usepackage{amsmath,amssymb,amsfonts}%
\usepackage{amsthm}%
\usepackage{mathrsfs}%
\usepackage[title]{appendix}%
\usepackage{xcolor}%
\usepackage{textcomp}%
\usepackage{manyfoot}%
\usepackage{booktabs}%
\usepackage{algorithm}%
\usepackage{algorithmicx}%
\usepackage{algpseudocode}%
\usepackage{listings}%
\usepackage{subcaption}%
\usepackage{tabularx} 
\usepackage{anyfontsize}

\theoremstyle{thmstyleone}%
\theoremstyle{thmstyletwo}%

\theoremstyle{thmstylethree}%

\begin{document}


\title[Article Title]{Enhanced Multi-Class Classification of Gastrointestinal Endoscopic Images with Interpretable Deep Learning Model}


\author[1]{\fnm{Astitva} \sur{Kamble}}
\author[1]{\fnm{Vani} \sur{Bandodkar}}

\author[1]{\fnm{Saakshi} \sur{Dharmadhikary}}
\author[1]{\fnm{Veena} \sur{Anand}}
\author[2]{\fnm{Pradyut} \sur{Kumar Sanki}}
\author[3]{\fnm{Mei} \sur{X. Wu}}
\author[1]{\fnm{Biswabandhu} \sur{Jana}}

\affil[1]{\orgname{ABV-Indian Institute of Information Technology and Management}, \orgaddress{\city{Gwalior, India}}}
\affil[2]{\orgname{SRM University}, \orgaddress{\city{AP, India}}}
\affil[3]{\orgname{Massachusetts General Hospital}, \orgaddress{Harvard Medical School, USA}}




\abstract{
Endoscopy serves as an essential procedure for evaluating the gastrointestinal (GI) tract and plays a pivotal role in identifying GI-related disorders. Recent advancements in deep learning have demonstrated substantial progress in detecting abnormalities through intricate models and data augmentation methods.
This research introduces a novel approach to enhance classification accuracy using 8,000 labeled endoscopic images from the Kvasir dataset, categorized into eight distinct classes. Leveraging EfficientNetB3 as the backbone, the proposed architecture eliminates reliance on data augmentation while preserving moderate model complexity. The model achieves a test accuracy of 94.25\%, alongside precision and recall of  94.29\% and 94.24\% respectively. Furthermore, Local Interpretable Model-agnostic Explanation (LIME) saliency maps are employed to enhance interpretability by defining critical regions in the images that influenced model predictions. Overall, this work highlights the importance of AI in advancing medical imaging by combining high classification accuracy with interpretability.}




\maketitle

\section{Introduction}\label{sec:intro}
Gastrointestinal (GI) diseases pose significant challenges to global healthcare systems, leading to considerable morbidity and mortality. Recent studies indicate that digestive disorders were responsible for about 2.56 million deaths and 88.99 million disability-adjusted life years (DALYs) worldwide in 2019, marking increases of 37.85\% and 23.47\%, respectively, since 1990 \cite{stat}. Disorders such as colorectal cancer, ulcers, gastritis, and polyps affect millions worldwide \cite{sung2021global}, emphasizing the need for early diagnosis and intervention to mitigate their impact.

Endoscopy has transformed the diagnostic and therapeutic landscape for GI disorders over the last two decades. It provides a minimally invasive approach to visualize the gastrointestinal tract, enabling the detection of abnormalities ranging from small polyps to precancerous lesions. Despite its utility, the subjective interpretation of endoscopic images often limits its accuracy, resulting in inconsistencies across diagnoses. To address these challenges, AI technologies, especially deep learning (DL), have demonstrated significant potential in advancing medical imaging applications \cite{8853349,chitnis2022brain,HE2020106539}. AI models, especially Convolutional neural networks (CNNs) have shown remarkable efficacy in identifying GI abnormalities through endoscopic images \cite{shichijo2017cnn}. Modern approaches such as transfer learning further enhance classification tasks, outperforming conventional machine learning models \cite{wang2022convolutional, MOHAPATRA2023101942}. 
A variety of datasets, such as those used in \cite{JIN2024106387} for endoscopic images, are available. Additionally, there are class-specific datasets, for example, the ones in \cite{TAS2021106959} for polyp detection. Video-based datasets, like those in \cite{LEE2024105637}, have also been explored to address temporal patterns and dynamic features in endoscopy videos.
This study uses the Kvasir-V2 dataset \cite{KVASIR}, a well-known repository of labeled GI endoscopic images that has become an essential tool for developing and evaluating such AI algorithms. Nezhad et al. \cite{nezhad2024GI} proposed a GI tract lesion classification model combining CNNs with a 2-D visualization method, while using the Kvasir-V1 dataset. Their approach outperforms traditional CNN-based models and utilizes endoscopic video frames for automatic lesion classification.

In recent years, AI-assisted computer-aided diagnostic (CAD) technologies have become enhanced tools for identifying subtle patterns and anomalies in endoscopic data \cite{hassan2020}. \textit{Ramzan et al.}
 \cite{Ramzan2021-CADx} developed a hybrid approach combining texture-based features with deep learning techniques, achieving 95.02\% accuracy on a subset of 4000 Kvasir dataset images. Similarly, studies focusing on early-stage conditions, such as esophagitis, highlight the importance of CNN-based frameworks for effective detection. \textit{Yoshiok et al.} \cite{yoshiok2023deeplearningmodelsmedicalimage} conducted a comparative analysis of CNN architectures, demonstrating superior F1-scores for GoogLeNet and high prediction confidence with MobileNetV3. GoogLeNet achieved approximately 84.6\% accuracy when classifying esophagitis and z-line images from the Kvasir dataset.
Öztürk and Özkaya \cite{ozturk2020gastrointestinal}  proposed an improved LSTM-based CNN for gastrointestinal tract classification. Their approach enhances CNN performance by integrating LSTM, demonstrating superior results over CNN+ANN and CNN+SVM classifiers. Another noteworthy contribution was made by \textit{Al-Otaibi et al.} \cite{rehman}, who explored transfer learning approaches applied to the Kvasir dataset, consisting of eight GI image categories. They employed data preprocessing and augmentation techniques to train various pretrained models. Their experiments yielded diverse accuracy levels - VGG16 (97.61\%), CNN (95.22\%), ResNet50 (96.61\%), Xception (77.94\%), and DenseNet121 (88.11\%). Among these, their proposed EfficientB1 model stood out, achieving 99.22\% accuracy. However, this dependency on data preprocessing and augmentation can raise concerns regarding assessing the model’s reliability and generalizability in real-life clinical scenarios..
\textit{Patel et al.} \cite{PATEL} discussed an explainable AI framework designed for gastrointestinal disease diagnosis within telesurgery healthcare systems. Their approach utilized ResNet50 to classify gastrointestinal images into eight distinct categories, achieving a validation accuracy of 92.52\%, slightly outperforming MobileNetV2, which attained 92.37\%. Despite these promising results, such accuracies may still fall short in scenarios demanding higher reliability during testing. To improve interpretability, their framework—TeleXGI—integrated Explainable Artificial Intelligence (XAI) frameworks, exemplified by Local Interpretable Model-agnostic Explanations (LIME), Integrated Gradients, Saliency Maps, and Grad-CAM visualizations, enabling deeper insights into model predictions. Padmavathi et al. \cite{padmavathi2023WCE} introduced a deep learning-based segmentation and classification approach for wireless capsule endoscopy (WCE) images, leveraging the Expectation-Maximization (EM) algorithm and DeepLab v3+ for feature extraction . 
\textit{Ahmed et al.} \cite{ahmed2023hybrid} explored hybrid models to interpret and analyze images obtained through endoscopy, emphasizing proactive identification of gastrointestinal diseases images. 
They employed pretrained models, including GoogLeNet (classification accuracy: 88.2\%), MobileNet (classification accuracy: 86.4\%), and DenseNet121 (classification accuracy: 85.3\%). Performance was further enhanced by hybrid strategies combining CNNs with FFNN (Feedforward Neural Networks)  and XGBoost, reaching a maximum accuracy of 97.25\%. These models utilized fused CNN features and gradient vector flow (GVF)-based segmentation techniques, demonstrating the potential of hybrid methods in improving diagnostic accuracy.

While complex models with a high number of parameters often yield superior performance, they also introduce challenges, including higher computational costs and the risk of overfitting. Enlarging neural network architectures does not always result in proportional performance gains, as optimization algorithms may struggle to effectively leverage the additional capacity, sometimes leading to underfitting \cite{dauphin2013bigneuralnetworkswaste}. Hybrid approaches, though effective, frequently demand extensive computational resources, including memory and training time, posing difficulties for deployment in resource-constrained settings \cite{hybrid_models_disadvantage}. Conversely, simpler architectures, while computationally efficient, may lack sufficient accuracy, limiting their utility in real-world applications.
Another challenge involves data augmentation techniques, which are often employed to expand training datasets artificially. The augmentation technique introduces dependencies and may compromise the model’s ability to handle raw, unprocessed data effectively \cite{Hüttenrauch2016}. In addition, manual tuning of augmentation parameters may  distort critical image features, leading to diagnostic errors, including false positives or negatives \cite{pattilachan2022criticalappraisaldataaugmentation}.

This study addresses the challenges posed by both excessively complex and overly simplistic model designs by introducing a balanced architecture optimized for gastrointestinal image analysis. EfficientNetB3 is used as the base model and custom layers are trained on top of it.
EfficientNetB3 provides an optimal balance of accuracy and computational efficiency, which makes appropriate for handling high-resolution medical images. By tuning the parameters and hyperparameters of the custom layers, the proposed method achieves high accuracy while avoiding computational overhead and overfitting risks associated with larger networks. Unlike many existing approaches, our model eliminates reliance on data augmentation, thereby improving robustness and adaptability to real-world datasets.
To further enhance transparency and trustworthiness, we integrate explainable AI (XAI) techniques, such as LIME, to highlight key regions influencing predictions. This interpretability allows clinicians to validate predictions, promoting actionable insights for medical decision-making. Our approach effectively balances computational efficiency, generalizability, and explainability, making it highly suitable for use in resource-constrained environments.

The subsequent sections of this paper are structured as follows: Section~\ref{sec:methods} outlines the methods, providing an overview about Kvasir dataset (Section~\ref{sec:dataset}), data preprocessing steps (Section~\ref{sec:preprocessing}), proposed model architecture (Section~\ref{sec:model}), and optimization techniques with hyperparameter configurations (Section~\ref{sec:optimization}). Section ~\ref{sec:training} explains the training procedure. Section~\ref{sec:results} presents the results, including performance metrics (Section~\ref{sec:performance}) and interpretability analysis (Section~\ref{sec:interpretability}). Finally, Section~\ref{sec:conclusion} concludes with a summary of findings and directions for future research.

\section{Methods}\label{sec:methods}

\subsection{Kvasir Dataset}\label{sec:dataset}
\begin{figure}[!htbp]
    \centering
    \scalebox{0.92}{
    \begin{tabular}{cccc}
    \includegraphics[width=90pt, height=85pt]{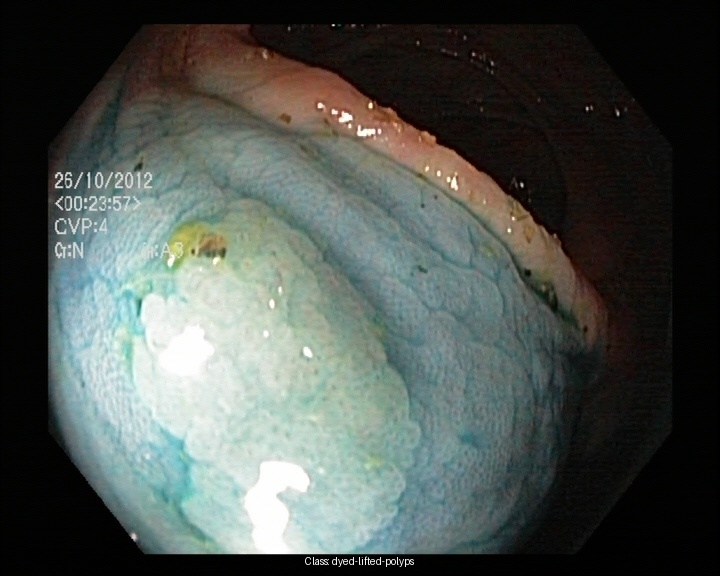} & 
    \includegraphics[width=90pt, height=85pt]{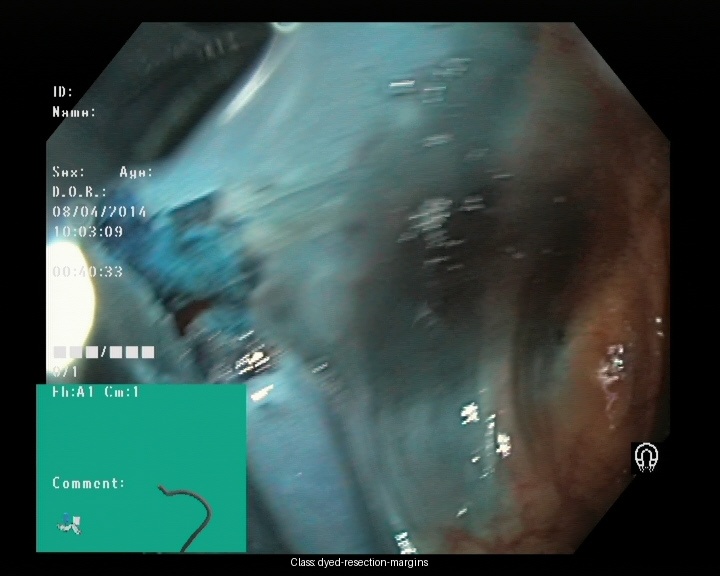} &
    \includegraphics[width=90pt, height=85pt]{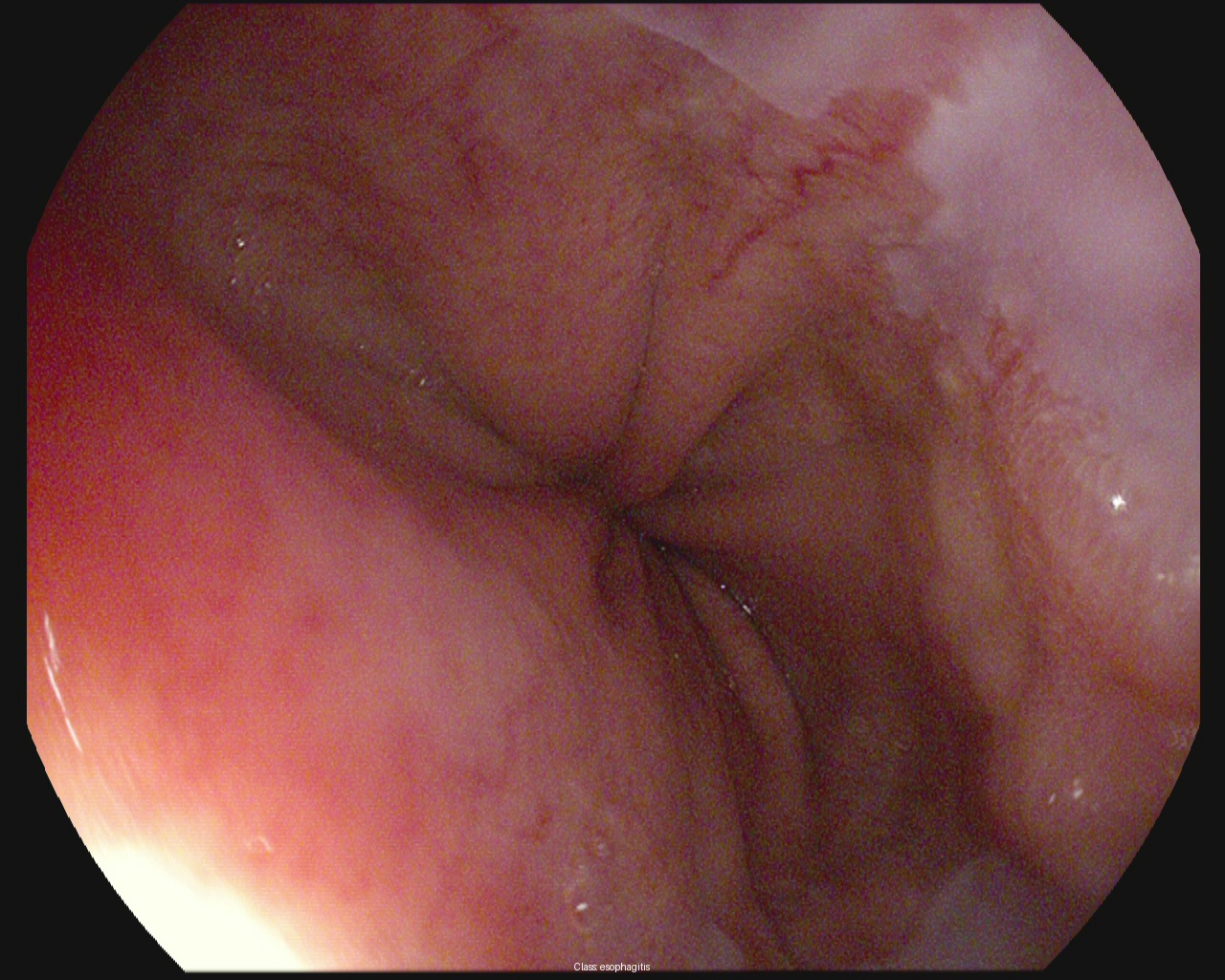} & 
    \includegraphics[width=90pt, height=85pt]{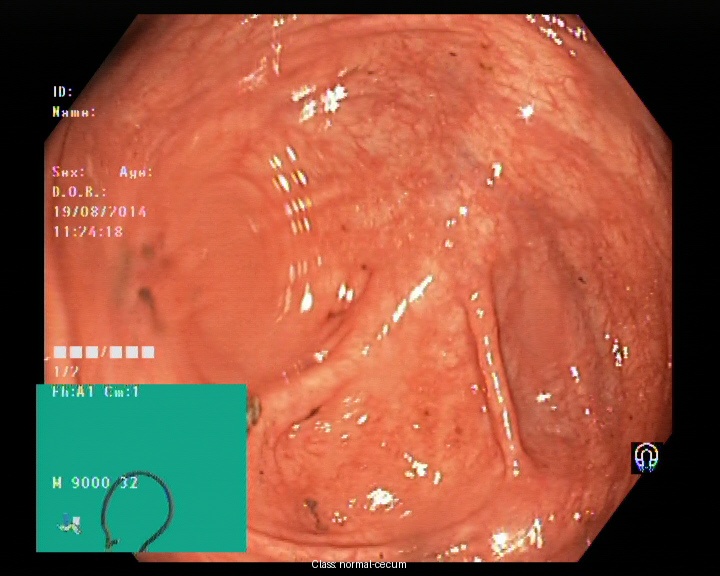} \\
    (a) & (b) & (c) & (d) \\

    \includegraphics[width=90pt, height=85pt]{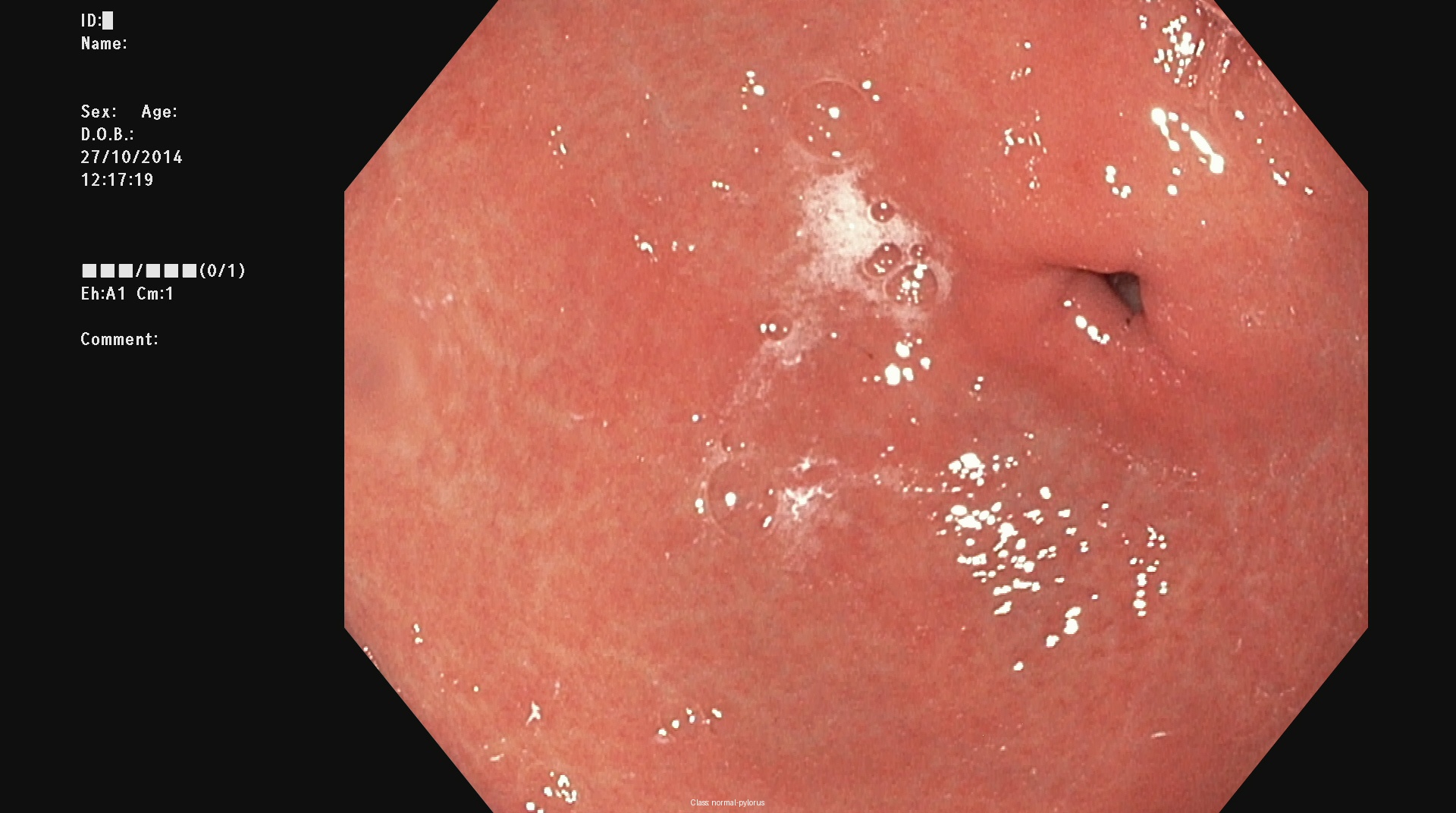} & 
    \includegraphics[width=90pt, height=85pt]{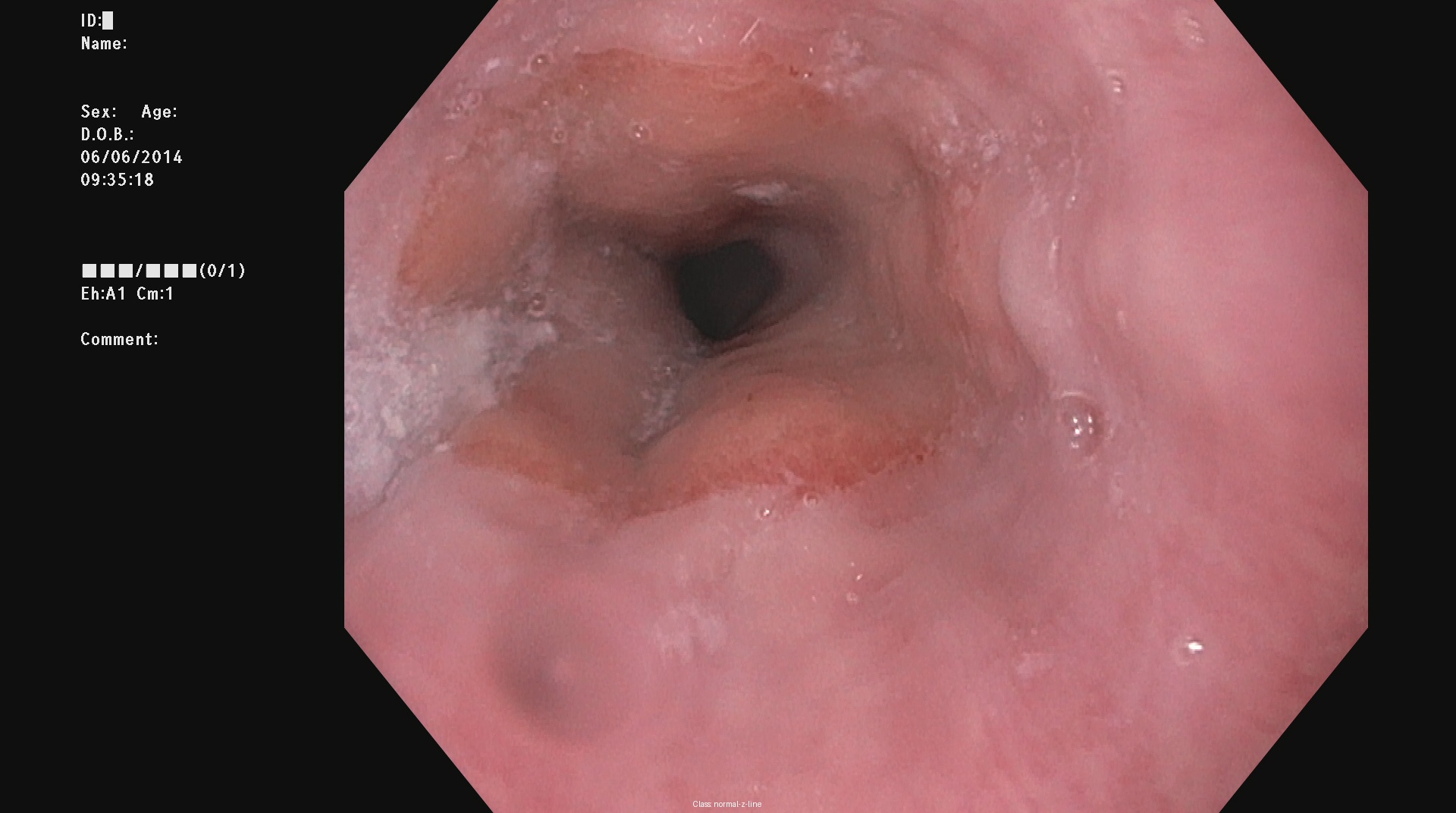} &
    \includegraphics[width=90pt, height=85pt]{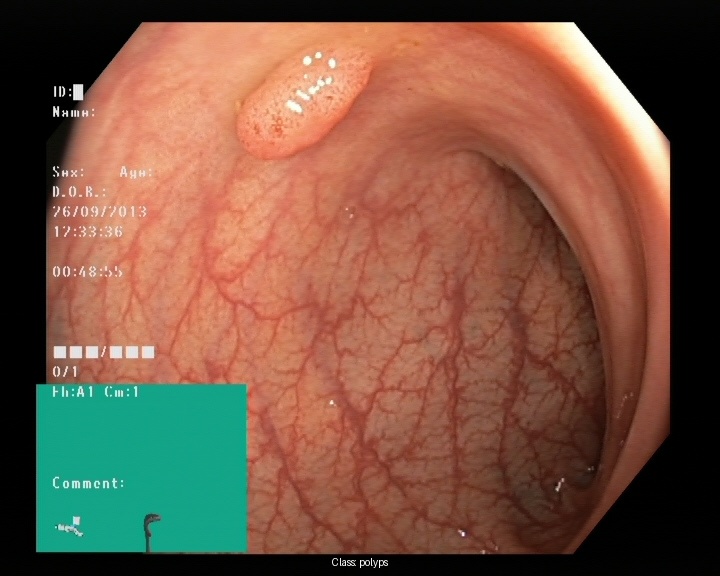} & 
    \includegraphics[width=90pt, height=85pt]{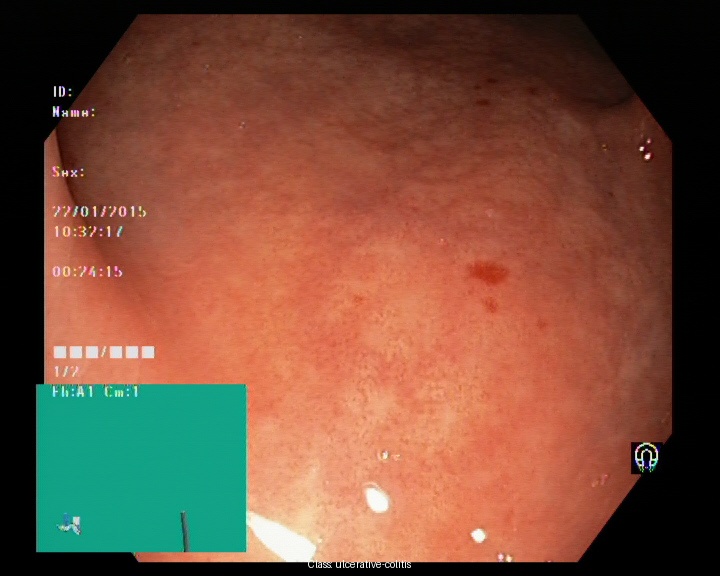} \\
    (e) & (f) & (g) & (h) \\
    \end{tabular}}
    \caption{\label{fig:fig1}Sample images from the Kvasir dataset, a widely used repository of GI endoscopic images designed for training and evaluating machine learning algorithms for automated diagnosis. Each labeled image corresponds to distinct GI findings, aiding in the development of diagnostic tools: (a) dyed-lifted polyps and (b) dyed resection margins—where staining highlights abnormalities for precise tissue removal; (c) esophagitis, characterized by inflamed esophageal tissue often associated with acid reflux; and (d), (e), and (f), showing normal anatomical landmarks such as the cecum, pylorus, and Z-line—essential for confirming complete visualization during endoscopy. Panels (g) and (h) highlight polyps with malignancy potential and ulcerative colitis, a chronic inflammatory condition requiring ongoing monitoring. The dataset's variety supports robust model development for AI-assisted diagnostics in gastroenterology.}
\end{figure}

Kvasir dataset serves as a widely recognized resource for research in automated gastrointestinal (GI) disease detection \cite{KVASIR}. It contains labeled endoscopic images representing a range of GI conditions, including both healthy and pathological cases. Its diversity in anatomical regions and pathological categories makes it a valuable benchmark for training machine learning models aimed at GI disease classification. The dataset has 8000 images belonging to 8 different classes namely dyed-lifted-polyps, dyed-resection-margins, esophagitis, normal-cecum, normal-pylorus, normal-z-line, polyps, ulcerative-colitis. Figure \ref{fig:fig1} depicts the sample images from the dataset. Despite its extensive coverage, the dataset poses challenges such as variations in lighting, differences in image quality, and visual similarities between certain categories, which may affect model performance and classification accuracy.

\subsection{Data Preprocessing}\label{sec:preprocessing}
The preprocessing of data is an important step in ensuring consistent input formatting for training, validation, and testing phases. To match the input requirements of the EfficientNetB3 model \cite{tan2020efficientnetrethinkingmodelscaling}, images are resized to \(224 \times 224\) pixels. To improve the model’s ability to generalize, horizontal flipping is applied during preprocessing to introduce variations in training samples. Additionally, input values are normalized using scalar preprocessing to standardize data across all phases.
The dataset is partitioned into three sets: training, validation, and testing, to assess the model’s performance, each organized as dataframes containing file paths and corresponding class labels. Batch processing is employed for efficiency, enabling simultaneous handling of multiple samples during training and testing.

\noindent \textbf{Image Dimensions and Format:}  
Each image is resized to \(224 \times 224\) pixels to meet the input requirements of the EfficientNetB3 architecture. Images are preprocessed in RGB format, utilizing three color channels (red, green, and blue) to preserve visual information critical for classification tasks.

\noindent \textbf{Custom Batch Size for Testing:}  
To optimize computational efficiency for testing, 64 is used as the size of the batch. This approach balances memory utilization and processing speed, ensuring smooth evaluation without resource constraints.

\subsection{Proposed Model Architecture}\label{sec:model}

\begin{figure}[ht]
    \centering
    \includegraphics[width=0.99\textwidth]{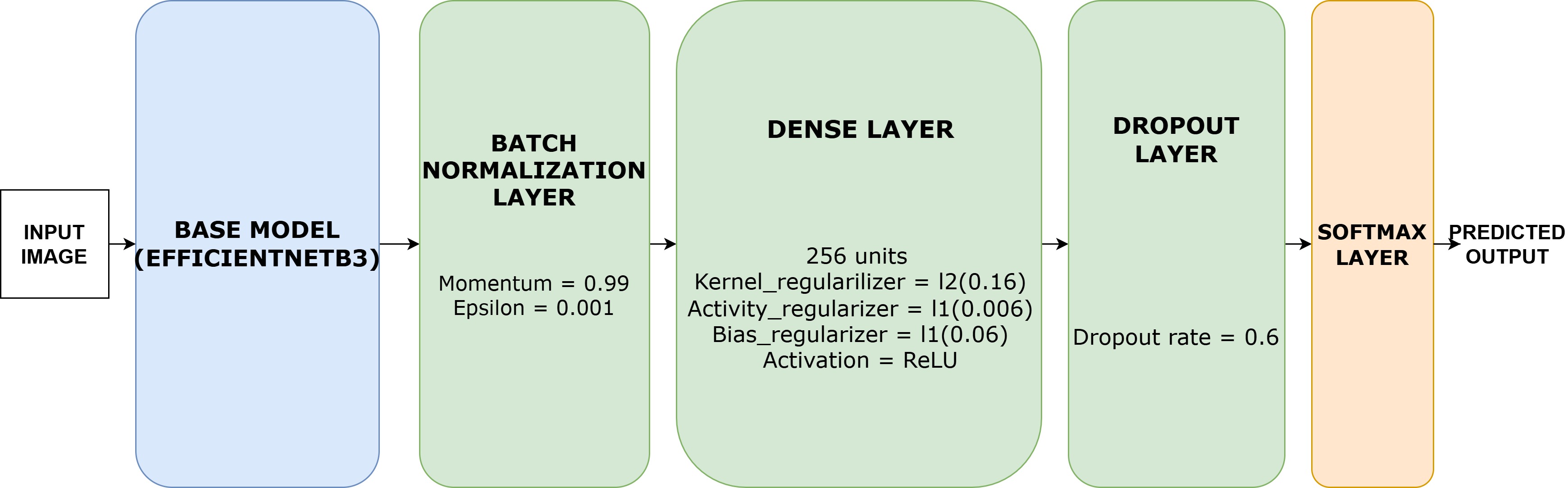}
    \caption{Block diagram representing the proposed model architecture. The design includes a base model (EfficientNetB3), additional custom layers, and a softmax layer at the output for predicting class probabilities.}
    \label{fig:model-architecture}
\end{figure}

Figure \ref{fig:model-architecture} presents the architecture of the proposed model , it integrates a pre-trained base model, followed by customized layers optimized with specific hyperparameters, and concludes with a softmax layer for class prediction. After evaluating multiple architectures, the EfficientNetB3 \cite{tan2020efficientnetrethinkingmodelscaling} model demonstrated the best performance for this task.

The first layer following the base model is the Batch Normalization \cite{ioffe2015batch} layer, which standardizes the inputs along the specified axis, typically corresponding to the channels in a feature map. It employs a momentum value of 0.99 to maintain the running mean and variance, along with an epsilon value of 0.001 for numerical stability. This normalization step accelerates training and improves generalization by reducing internal covariate shifts.


Subsequently, a Dense layer with 256 neurons is incorporated to capture complex patterns extracted by the base model. The ReLU  \cite{agarap2018deep} activation function is applied to introduce non-linearity, along with regularization techniques to mitigate overfitting and enhance generalization.

The regularization methods include penalties based on L1 \cite{tibshirani1996lasso} and L2 \cite{hoerl1970ridge}. L1  imposes a penalty that is proportional to absolute magnitude of the weights, encouraging sparsity by effectively setting less significant parameters to zero:

\[
\text{L1 Penalty} = \lambda \sum |w_i|
\]

where \( \lambda \) is the regularization factor and \( w_i \) represents the model parameters. L2 regularization penalizes the squared values of weights to prevent excessively large coefficients:

\[
\text{L2 Penalty} = \lambda \sum w_i^2
\]

An L2 penalty of 0.16 is applied to weight values, while an L1 penalty of 0.006 is imposed on output activations, encouraging sparsity. Additionally, an L1 penalty of 0.06 is applied to biases, helping reduce their magnitudes and dependency on constant shifts. These techniques collectively balance model complexity and generalization performance.

Following the \texttt{Dense} layer, to minimize overfitting and increase generalization, a \texttt{Dropout} \cite{srivastava2014dropout} layer is applied. During training, a randomly selected fraction of the input units is dropped by setting them to zero, forcing network to learn distributed representations. For this architecture, With a dropout rate of 0.6, 60\% of the units are deactivated at each update cycle. The output values are scaled by a factor of \( \frac{1}{1 - \text{rate}} \) to maintain activation sums. The dropout process can be mathematically defined as:

\[
y_i = 
\begin{cases} 
    0 & \text{if } r_i < \text{rate} \\
    \frac{x_i}{1 - \text{rate}} & \text{otherwise}
\end{cases}
\]

where \( r_i \) is a random value sampled uniformly between 0 and 1, \( x_i \) is the input, and \( y_i \) is the output after applying dropout. This technique encourages the network to rely on multiple neurons rather than single dominant units, thereby improving robustness.

A \texttt{Dense} layer forms the final layer, with neurons matching the number of output classes. The \texttt{softmax} activation function is used to transform logits into probabilities for each class in multi-class classification. A softmax function for class \( i \) is defined as:

\[
P(y_i) = \frac{\exp(z_i)}{\sum_{j=1}^{N} \exp(z_j)}
\]

where \( z_i \) corresponds to logit of class \( i \), \( N \) denotes total number of classes, and \( P(y_i) \) is predicted probability. The outputs are normalized to ensure they sum to one, providing meaningful class probabilities for predictions.

\subsection{Optimization Techniques and Hyperparameter Configuration}\label{sec:optimization}

To enhance the efficiency and effectiveness of training, this work employs a combination of custom callbacks and carefully selected hyperparameters. These mechanisms ensure optimal performance while preventing overfitting:

\textbf{Dynamic Learning Rate Scheduling:} The learning rate is adjusted dynamically by reducing it by a predefined factor of 0.5 when performance metrics, such as validation loss, stagnate for a specified patience period of 3 epochs. This gradual adjustment facilitates smooth convergence as training progresses.

\textbf{Early Stopping Mechanism:} Training is halted after 5 epochs of no significant improvement, avoiding unnecessary computations and saving time. This mechanism is complemented by threshold monitoring to determine when to prioritize different metrics.

\textbf{Threshold Monitoring and Metric Prioritization:} Metrics such as training accuracy and validation loss are monitored dynamically. Early stages emphasize accuracy, while later stages prioritize loss if accuracy surpasses a defined threshold of 0.9. This ensures focus on the most relevant aspects of model performance at different stages of training.

\textbf{Batch Processing and Epoch Control:} Training is conducted with a batch size of 64, balancing computational efficiency and memory utilization. A total of 15 epochs is used, as determined through experimentation, providing sufficient training while minimizing the risk of overfitting.

\textbf{Manual Control for Adaptive Training:} Periodic prompts enable user intervention, allowing adjustments to training duration or termination based on intermediate observations. This flexibility ensures adaptability to unexpected patterns in data behavior.

\textbf{Performance Tracking and Model Preservation:} The best-performing model configuration is continuously monitored and saved, preserving optimal weights even if later epochs degrade performance. This approach ensures that the final model retains its peak predictive capability.

\textbf{Time Monitoring and Resource Management:} Epoch durations and cumulative training times are tracked to evaluate computational efficiency. These insights guide future optimizations and resource allocation for similar tasks.

\subsection{Model Compilation and Training}\label{sec:training}

After defining the architecture, as outlined in the previous section, and setting the required hyperparameters, The \texttt{Adamax} optimizer \cite{kingma2017adammethodstochasticoptimization} is used to compile the model with a learning rate set to 0.001. Adamax, an extension of the Adam optimizer, utilizes the infinity norm to scale gradient updates, offering enhanced stability when dealing with sparse gradients or noisy data. It modifies the learning rate for each parameter in real-time, depending on its gradient history, making it particularly effective in cases involving large or sparse gradients.
The parameter update mechanism employed by the Adamax optimizer can be expressed as:

\begin{equation}
    \theta_t = \theta_{t-1} - \frac{\eta}{\hat{v}_t^{\infty} + \epsilon} \cdot \hat{m}_t
\end{equation}

where:
\begin{itemize}
    \item \( \theta_t \) denotes the parameter value at time step \( t \),
    \item \( \eta \) represents the learning rate,
    \item \( \hat{m}_t \) corresponds to the bias-corrected first-moment estimate, similar to that in the Adam optimizer,
    \item \( \hat{v}_t^{\infty} \) denotes the infinity norm of the second-moment estimate, which is computed as:
\end{itemize}

\begin{equation}
    \hat{v}_t^{\infty} = \max_{i} \left( |v_{t-1}^i| \right)
\end{equation}

Here, \( \hat{v}_t^{\infty} \) captures the maximum absolute value of the second-moment estimates across all parameters. 

\begin{itemize}
    \item \( \epsilon \) is a small constant introduced to avoid division by zero.
\end{itemize}

This formulation ensures stable parameter updates, particularly in scenarios involving large gradient values or sparse features.

This study employs \texttt{categorical\_crossentropy} as the loss function, making it well-suited for multi-class classification tasks where each input is categorized into one of several classes. It evaluates the disparity between the predicted and true class probabilities, penalizing any deviations from the correct labels. For an individual sample, the categorical crossentropy loss is calculated as:
\begin{equation}
    L = - \sum_{i=1}^{C} y_i \log(p_i)
\end{equation}

where:
\begin{itemize}
    \item \( C \) denotes the total number of classes,
    \item \( y_i \) represents the true probability distribution (1 for the correct class, 0 otherwise),
    \item \( p_i \) corresponds to the predicted probability for class \( i \).
\end{itemize}

The loss function is designed to enhance probabilities for the true class while reducing the likelihoods associated with incorrect classes, leading to better classification performance.
The training process spans 15 epochs and utilizes the prepared training dataset. Throughout training, performance is dynamically monitored, and adjustments are made to optimize learning efficiency and prevent overfitting. These adjustments include refining the learning rate and tracking progress against validation metrics.
At the end of each epoch, validation data is analyzed to evaluate the model’s ability to generalize to unseen samples. Since the dataset is pre-organized to maintain consistency, data shuffling is excluded during training. This ensures reproducibility and preserves evaluation integrity.

\section{Results and Discussions}\label{sec:results}
\subsection{Model Performance and Comparisons}\label{sec:performance}

The experimental evaluation of the proposed model is conducted using both training and testing datasets. The training-testing split was 80\% for training and 20\% for validation and testing. The model underwent training for 15 epochs on a dataset comprising 6,400 images distributed across 8 distinct classes. A validation set consisting of 800 images from the same 8 classes is used to monitor performance during training. Key metrics, including accuracy, validation accuracy, loss, and validation loss, are tracked throughout the training process.

\begin{figure}[ht!]
    \centering
    \begin{subfigure}[t]{0.48\textwidth}
        \centering
        \includegraphics[width=\textwidth, keepaspectratio]{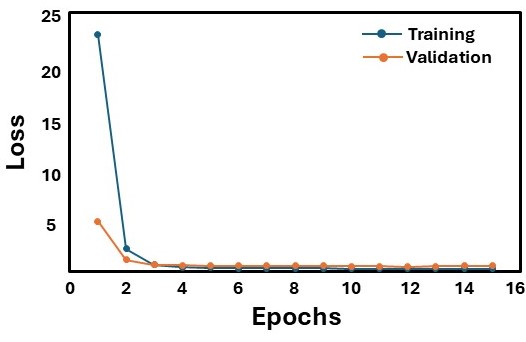}
        \caption{Loss vs Epochs.}
        \label{fig:loss_vs_epochs}
    \end{subfigure}
    \hfill
    \begin{subfigure}[t]{0.51\textwidth}
        \centering
        \includegraphics[width=\textwidth, keepaspectratio]{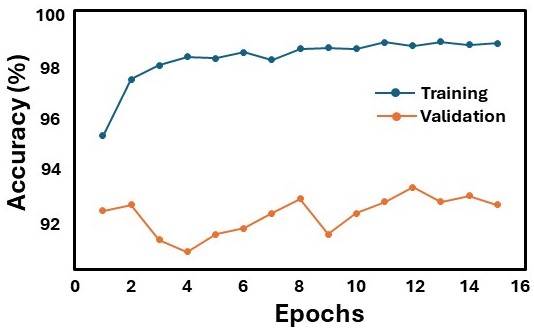}
        \caption{Accuracy vs Epochs.}
        \label{fig:accuracy_vs_epochs}
    \end{subfigure}
    \caption{Training and Validation Metrics. (a) Loss reduction trend during training. (b) Accuracy progression during training.}
    \label{fig:metrics_plots}
\end{figure}

Figure \ref{fig:loss_vs_epochs} presents the Loss vs. Epochs plot, illustrating a consistent reduction in loss values over successive epochs. This trend highlights the model’s ability to learn effectively. Similarly, Figure \ref{fig:accuracy_vs_epochs} depicts the Accuracy vs. Epochs plot, with a peak training accuracy of 99.61\%, the model attained a validation accuracy of approximately 93.50\%. These results demonstrate the model's strong capacity for generalization.

The model undergoes performance evaluation on a test set after training. The test set contains 800 unseen images from the same 8 classes, ensuring a robust assessment of its real-world applicability. To evaluate performance, various metrics are employed, including accuracy, recall, precision, specificity, F1-score \cite{grossman2004information}, and inference time.
A comparative analysis, is presented in Table \ref{tab1}, of the proposed model against other architectures tested during development. The evaluation criteria included testing accuracy, parameter complexity, and computational efficiency. Simpler models, such as MobileNetV3 \cite{howard2019mobilenetv3}, which has approximately 3.2 million parameters, achieved a testing accuracy of 90.62\%. On the other hand, more complex architectures, such as ResNet152 \cite{he2016deep}, with around 58.8 million parameters, attained a higher accuracy of 93.97\%.
Additional models, including InceptionResNetV2 \cite{szegedy2016inception} and its extended version, InceptionResNetV2L, comprising 54.6 million and 54.7 million parameters, achieved accuracies of 92.25\% and 92.87\%, respectively. However, their computational requirements posed challenges in terms of efficiency. Comparatively, VGG19 yielded a lower accuracy of 65.37\%, highlighting its limitations despite its simplicity.
Other notable architectures include DenseNet201 \cite{huang2017densely}, achieving 93.62\%, and Xception \cite{chollet2017xception}, reaching 92.75\%. Hybrid approaches, such as combinations of EfficientNetB3 and DenseNet201, demonstrated promising results with a testing accuracy of 93.88\%. However, these hybrid models required higher inference times, approximately 7.2 seconds, compared to an average of 3.2 seconds for other models.
The proposed model, which uses EfficientNetB3 as the base architecture, achieved a well-balanced approach to accuracy and computational efficiency. It recorded a testing accuracy of 94.25\%, with 11.1 million parameters and an average inference time of approximately 3.2 seconds. These results emphasize the model's superior performance, combining high classification accuracy with efficient computation, making it suitable for real-world applications.

\begin{table}[htbp]
    \centering
    \tiny
    \caption{Comparison of proposed method and other models on testing dataset}
    \addtolength{\tabcolsep}{-4pt}
    \renewcommand{\arraystretch}{2}
    \begin{tabular}{lccccccc}
        \hline
        \textbf{Methods} & \textbf{Accuracy(\%)} & \textbf{Recall(\%)} & \textbf{Precision(\%)} & \textbf{F1-Score(\%)}  & \textbf{Specificity(\%)} & \textbf{Parameters} & \textbf{Test Time} \\
        \hline
        EffNetB3+DenseNet201      & 93.88 & 93.88 & 93.90 & 93.88 & 99.12 & 29.6M & 7.2s \\
        DenseNet201 \cite{huang2017densely}     & 93.62 & 93.62 & 93.84 & 93.63 & 99.08 & 18.5M & 3.2s \\
        InceptionResNetV2 & 92.25 & 92.25 & 92.51 & 92.21 & 98.89 & 54.6M & 3.5s \\
        ResNet152 \cite{he2016deep}       &  93.97  & 93.38 & 93.64 & 93.40 & 99.05 & 58.8M & 3.3s \\
        MobileNetV3 \cite{howard2019mobilenetv3}     & 90.62   & 90.62 & 91.43 & 90.49 & 98.66 & 3.2M & 3.2s \\
        InceptionResNetV2L\cite{szegedy2016inception} & 92.87   & 91.00 & 91.23 & 91.03 & 98.71 & 54.7M & 3.3s \\
        Xception \cite{chollet2017xception}        & 92.75   & 92.75 & 93.04 & 92.67 & 98.96 & 21.3M & 3.2s \\
        VGG19            & 65.37   & 65.38 & 65.34 & 60.89 & 95.05 & 20.15M & 3.3s \\
        \textbf{EfficientNetB3}   & \textbf{94.25}   & \textbf{94.24} & \textbf{94.29} & \textbf{94.29} & \textbf{99.18} & \textbf{11.1M} & \textbf{3.5s} \\
        \hline
    \end{tabular}   
    \label{tab1}
\end{table}

\begin{figure}[ht!]
    \centering
    \includegraphics[width=1\textwidth, keepaspectratio]{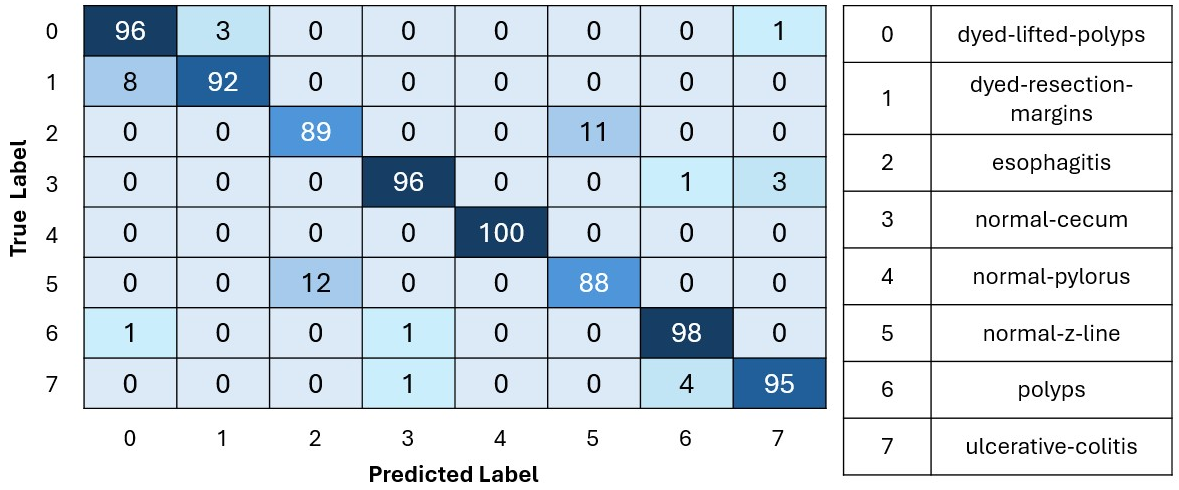}
    \caption{Confusion matrices for the proposed model using the test set.}
    \label{confusion_matrix}
\end{figure}

Figure \ref{confusion_matrix} illustrates the confusion matrix generated during the testing phase. This visualization provides a detailed overview of the model's predictions across all classes, emphasizing its ability to correctly classify most instances while reducing misclassifications. The high precision and recall values evident in the confusion matrix underscore the model's discriminative power and robustness across all classes. Furthermore, the matrix provides valuable insights into class-specific performance, enabling the identification of areas for potential improvement in future iterations of the model.

Additionally, a comparison of the proposed model with several recent models from the literature is shown in Table \ref{tab2}. In \cite{rehman}, the EfficientNet architecture was tested on two different datasets, achieving an accuracy of 99.22\%. However, this reported accuracy was likely obtained from the entire dataset, including data that may have been seen during training, potentially inflating the reported performance metrics. The study also employed data augmentation techniques to enhance model performance. Other studies using data augmentation include \cite{PATEL}, which achieved an accuracy of 98.8\%, \cite{Ayşe} with an accuracy of 93.37\%, and \cite{s23063176}, which reported an accuracy of 93.46\%.
In contrast, our model achieved a testing accuracy of 94.25\% without utilizing any data augmentation techniques, demonstrating its ability to generalize effectively to real-world images. Another recent study \cite{Gunasekaran} achieved a higher accuracy of 95.00\% using hybrid models, which involved a significantly increased number of parameters.
These comparisons emphasize that our model, even without data augmentation and tested exclusively on unseen images, outperforms many existing models by a substantial margin. This reinforces effectiveness of the model, robustness, and practical relevance in real-world applications, where the absence of augmentation maintains robust performance.

\begin{table}[htbp] 
    \centering
    \tiny
    \caption{Comparison of proposed method with other methods}
    \addtolength{\tabcolsep}{-4pt}
    \renewcommand{\arraystretch}{2}
    \begin{tabular}{|l|c|c|c|c|c|c|c|c|}
         \hline
        Models & Accuracy & Recall & Precision &  Data Augmentation \\
         \hline
        EfficientNetB3 (Proposed)  & 94.25   & 94.24 & 94.29 &   No \\
        EfficientNet \cite{rehman} (2024)  & 99.22   & 99.22 & 99.22 &  Yes \\
        TeleXGI	ResNet50 \cite{PATEL} (2024)  & 98.8   & 98.87 & 98.93 &  Yes \\
        Spatial-attention ConvMixer \cite{Ayşe} (2024)  & 93.37   & 93.37 & 93.66 &  Yes \\
        ResNet and Grad–CAM \cite{s23063176} (2023)      & 93.46 & - & - &  Yes\\
        DenseNet201-InceptionV3-ResNet50 \cite{Gunasekaran} (2023)      & 95.00 & 94.66 & 94.33 &  Yes\\
        CapsNet \cite{AYIDZOE} (2022)      & 80.57 & - & 82.07 &  No\\
        \hline
    \end{tabular}   
    \label{tab2}
\end{table}




\subsection{Explainability Analysis (XAI)}\label{sec:interpretability}

In this work, we employ LIME \cite{DBLP:journals/corr/RibeiroSG16, 9233366} to provide interpretable insights into our endoscopic classification model based on the EfficientNet-B3 architecture. 
LIME provides insights into how AI interprets images, helping clinicians understand the impact of visual features on the AI's classification focus.
 
LIME operates by creating local approximations to explain complex model decisions.For an image classifier \( f \), LIME provides an explanation by locally approximating \( f \) with a simple, interpretable model \( m \) near the instance being explained. This local approximation can be formulated as:

\begin{equation}
    \phi(x) = \arg \min_{m \in M} \, \mathcal{D}(f, m, \rho_x) + \mathcal{C}(m)
\end{equation}

where:
\begin{itemize}
    \item \( \phi(x) \) denotes the explanation for the input \( x \).
    \item \( \mathcal{D}(f, m, \rho_x) \) quantifies how well \( m \) approximates \( f \) in the neighborhood defined by \( \rho_x \).
    \item \( \mathcal{C}(m) \) represents a regularization term that penalizes the complexity of the model \( m \).
    \item \( M \) refers to the set of interpretable models, often linear or simple decision trees.
    \item \( \rho_x \) defines the local neighborhood around the instance \( x \) used for the approximation.
\end{itemize}

\begin{figure}[!htbp]
    \centering
    \scalebox{0.92}{
    \begin{tabular}{cccc}
      \includegraphics[width=90pt, height=85pt]{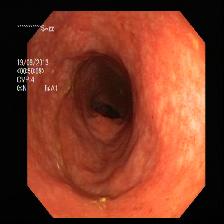} & \includegraphics[width=90pt, height=85pt]{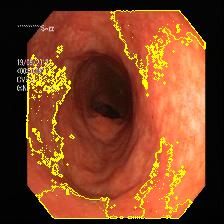} &
      \includegraphics[width=90pt, height=85pt]{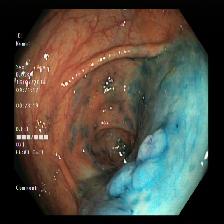} & \includegraphics[width=90pt, height=85pt]{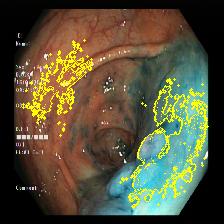}\\
      (a) & (b) & (c) & (d)\\

      \includegraphics[width=90pt, height=85pt]{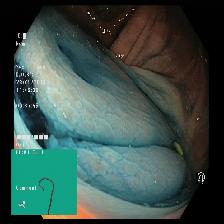} & \includegraphics[width=90pt, height=85pt]{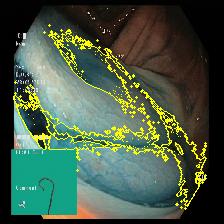} &
      \includegraphics[width=90pt, height=85pt]{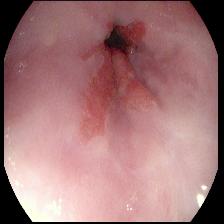} & 
      \includegraphics[width=90pt, height=85pt]{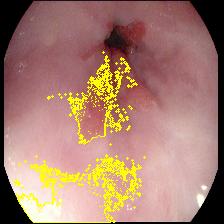}\\
      (e) & (f) & (g) & (h)\\

    \includegraphics[width=90pt, height=85pt]{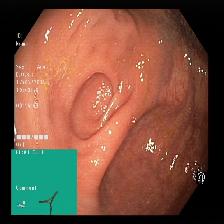} & \includegraphics[width=90pt, height=85pt]{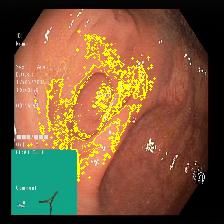} &
      \includegraphics[width=90pt, height=85pt]{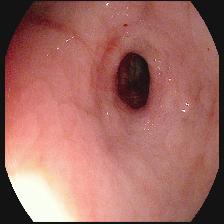} & \includegraphics[width=90pt, height=85pt]{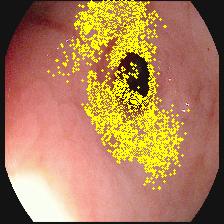}\\
      (i) & (j) & (k) & (l)\\

          \includegraphics[width=90pt, height=85pt]{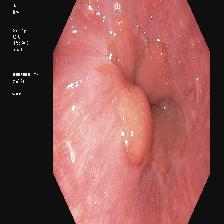} & \includegraphics[width=90pt, height=85pt]{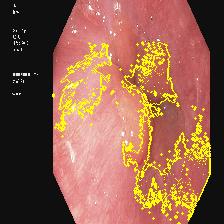} &
      \includegraphics[width=90pt, height=85pt]{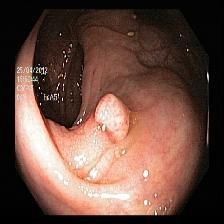} & \includegraphics[width=90pt, height=85pt]{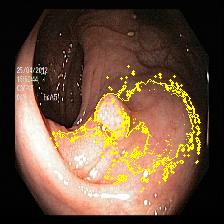}\\
      (m) & (n) & (o) & (p)\\
    \end{tabular}}
  \caption{\label{fig:label}This illustrates saliency maps generated using LIME for predictions on endoscopy images. The figure pairs the original endoscopic images (a, c, e, g, i, k, m, o) with their corresponding LIME-based explanation maps (b, d, f, h, j, l, n, p). The yellow overlays identify regions that significantly influence the model’s predictions, offering insights into its decision-making process. These visualizations enhance interpretability by highlighting the model’s focus areas and its potential to aid in diagnosing gastrointestinal conditions.}
\end{figure}

Figure \ref{fig:label} showcases paired examples of raw and LIME-explained images across various classes, providing direct comparisons and highlighting the influence of visual enhancements on the interpretability of the model. 

\subsection*{LIME Parameters}

For this study, the parameters of LIME are carefully configured to provide meaningful and interpretable insights into the classification decisions made by our endoscopic image classification model. To generate robust local explanations, the number of samples parameter is set to 1000, ensuring the creation of a sufficient number of perturbed samples around the input instance for accurate approximation.

To highlight the most informative image segments, the positive contributions only parameter is set to True, focusing solely on those superpixels that contributed positively to the predicted label. The number of features parameter is set to 5, limiting the explanation to the five most critical superpixels for interpretability. Additionally, the color to hide irrelevant regions parameter is assigned a value of 0, using a uniform black color to obscure non-relevant regions in the image during the explanation process. Finally, the minimum weight threshold parameter is set to 0, ensuring that all superpixels with weights above zero are included in the explanation.

The findings confirm LIME’s effectiveness in shedding light on the model’s decision-making process for both standard and dyed endoscopic images. With LIME generating visual explanations, clinicians can better understand the regions that influence the model’s predictions, thereby gaining insight into how visual enhancements guide the AI’s focus.

\section{Conclusion}\label{sec:conclusion}

This study introduces a novel model architecture based on EfficientNet B3 to enhance classification accuracy for gastrointestinal disease diagnosis using the Kvasir dataset. The proposed model achieves an accuracy of 94.25\% on unseen testing data, showcasing its effectiveness in practical scenarios. Its moderate complexity ensures a trade-off between performance and computational efficiency, making it well-suited for deployment in environments with limited resources. Notably, the model eliminates the need for data augmentation techniques, improving robustness and generalizability to real-world datasets. This approach can mitigate the risk of feature distortion, reduce manual fine-tuning, and address limitations observed in augmentation-based methods. Additionally, the application of XAI techniques, particularly LIME, yields valuable understanding of the model’s decisions. This interpretability helps clinicians understand the influence of visual features, enhancing trust and usability in diagnostic workflows. However, this study is limited to eight disease categories in the Kvasir dataset, and further evaluation on datasets with greater class diversity is necessary.
Additionally, the model has been validated on static image data, and its applicability to video-based datasets should be explored to handle temporal patterns and dynamic features in endoscopy videos. Another area for future work involves integrating the model with hardware systems, such as real-time diagnostic tools, to enable seamless deployment in clinical settings.
Testing the model across diverse clinical environments will ensure broader applicability and support clinical decision-making in gastrointestinal endoscopy.

\backmatter



\section*{Availability of the Data}
The publicly available Kvasir dataset used in this study can be accessed through the reference \cite{KVASIR}.  

\section*{Acknowledgments}
The authors express their thanks to FIG and CPDA of ABV-IIITM Gwalior, India for their valuable support.

\section*{Funding}
The authors received no specifc funding for this study.

\section*{Ethics Declarations}
Not applicable
\section*{Competing Interests}
The authors declare that they have no competing interests.

\section*{Consent to Participate}
All authors provide their consent to participante in this study. 

\section*{Consent to Publish}
All authors provide their consent for the publication of this manuscript.

\section*{Authors' Contributions}
Astitva and Vani were responsible for software implementation, and performing data analysis. Saakshi contributed to manuscript writing and literature review. Veena helped in conducting experiments. Pradyut assisted in result interpretation. Mei provided critical revisions and helped refine study objectives. Biswa supervised the entire research.




\bibliography{sn-article}

\end{document}